%% file: main_ACSOS_101719.tex
\documentclass[conference]{IEEEtran}
\IEEEoverridecommandlockouts
\usepackage{cite}
\usepackage{amsmath,amssymb,amsfonts}
\usepackage{algorithmic}
\usepackage{graphicx}
\usepackage{textcomp}
\usepackage{xcolor}
\usepackage[toc,abbreviations]{glossaries-extra}
\input{glossaries}

\usepackage{tabularx,ragged2e}
\usepackage{xltabular}
\newcolumntype{C}{>{\Centering\arraybackslash}X} 
\usepackage[caption=false]{subfig}
\usepackage[font=normalsize,labelfont={bf},justification=centering]{caption}
\renewcommand{\thetable}{\arabic{table}}
\def\tablename{Table}
\usepackage{adjustbox}
\usepackage{csquotes}
\usepackage{varwidth}
\usepackage{multirow}
\usepackage{booktabs}

\usepackage{multibib}

\def\BibTeX{{\rm B\kern-.05em{\sc i\kern-.025em b}\kern-.08em
    T\kern-.1667em\lower.7ex\hbox{E}\kern-.125emX}}
\begin{document}

\title{Comparing Generative Chatbots Based on Process Requirements: A Case Study}

\author{
	\IEEEauthorblockN{Luis Fernando Lins, Nathalia Nascimento, Paulo Alencar, Toacy Oliveira, Donald Cowan}
	\IEEEauthorblockA{\textit{David R. Cheriton School of Computer Science} \\
		\textit{University of Waterloo (UW)}\\
		Waterloo, Canada \\
		\{lflinsdossantos, nmoraesd, palencar, toliveira, dcowan\}@uwaterloo.ca}
}

\IEEEoverridecommandlockouts
\IEEEpubid{\makebox[\columnwidth]{979-8-3503-2445-7/23/\$31.00~\copyright2023 IEEE \hfill} \hspace{\columnsep}\makebox[\columnwidth]{ }}

\maketitle

\IEEEpubidadjcol

\begin{abstract}
Business processes are commonly represented by modelling languages, such as 
\gls{epc}, \gls{yawl}, and the most popular standard notation for modelling business processes, the \gls{bpmn}. Most recently, chatbots, programs that allow users
to interact with a machine using natural language, have been increasingly used for business process execution support. A recent category of chatbots worth mentioning is generative-based chatbots, powered by \glspl{llm} such as OpenAI's \gls{gpt} model and Google's \gls{palm}, which are trained on billions of parameters and support conversational intelligence. However, it is not clear whether generative-based chatbots are able to understand and meet the requirements of constructs such as those provided by \gls{bpmn} for process execution support. This paper presents a case study to compare the performance of prominent generative models, \gls{gpt} and \gls{palm}, in the context of process
execution support. The research sheds light into the challenging problem of using conversational approaches supported by generative chatbots as a means to  understand process-aware modelling notations and support users to execute their tasks. 

\end{abstract}


\section{Introduction}
Processes are pervasive in personal and organizational settings. They can support a wide variety of tasks, ranging from scheduling a trip \cite{bpm-book} to hiring new employees or admitting students into a university program \cite{dumas2018fundamentals}. For this reason, organizations are increasingly relying on automated processes to improve efficiency and accuracy in the execution of their tasks. 

In many cases, processes require people to perform tasks manually. These processes are known as human workflows \cite{dang2008human}. Although mechanisms have been proposed to visualize, control, and manage processes, performing tasks in a process involving numerous human decisions can be more complicated than it seems. First, tasks may have either a strict or a loose order, and this might leave participants confused in complex scenarios. Second, dependencies between tasks can be challenging to track, in which case, not being aware of the information that is required \textit{a priori}, people may start executing tasks they are not able to finish properly. Third, processes that rely heavily on humans may differ based on their context or require access to ancillary documentation so that errors are avoided \cite{dijkman2012managing}. Owing to these challenges, process participants should have a solution to guide them through process instances. 

One possible solution to help guide users through business processes are chatbots, also known as conversational agents. Chatbots are an emerging technology that has become ever so present in various domains, from healthcare to customer service \cite{luo2022critical}. This technology’s main selling point is that chatbots are able to talk to humans, understand their needs, and respond in a coherent way, while also being available 24/7.  Many businesses and academic projects have been using chatbots as a valuable tool, but there are still some challenges in the chatbot field that remain to be solved in an effective way. Some of these challenges include personalized responses based on user context, analysis of customer behaviour from interaction history, sentiment and emotion analysis, and negotiation without human intervention \cite{cheng2021exploring} \cite{shumanov2021making}.

On the other hand, chatbots can be very unpredictable, deciding conversation paths by itself that not always correspond to the business's intention, especially when the chatbot is machine-learning-based \cite{wang2018living}. Therefore, another challenge in the development of chatbots is creating one that closely follows the business requirements. It is not clear whether generative-based chatbots are able to understand and meet the requirements of constructs such as those provided by BMPM for process execution support. 
For this reason, this paper presents a case study to compare the performance of prominent generative models, \gls{gpt} and \gls{palm}, in the context of process
execution support. The research sheds light into the challenging problem of using conversational approaches supported by generative chatbots as a means to  understand process modelling notations and support users to execute their tasks. 

Our strategy involves two steps. First we produce the \gls{xml} file related to a specific case study (e.g., travel planning) using a BPMN-modelling application, in this case, Camunda Modeler. Second, we provide the chatbot with the entire \gls{xml} file related to a \gls{bpmn} model, and let it interpret it on its own. We will send the \gls{xml} file related to one \gls{bpmn} process model to two different \gls{llm} chatbots (\gls{gpt}-3.5 and \gls{palm} 2), in an attempt to see how well each model follows the original process model, based on a set of evaluation questions that will be described in this section. 

The paper is structured as follows. Section \ref{background} provides some research background and related work. Section \ref{generating-xml} describes the generation of \gls{xml} of the \gls{bpmn} process for our case study. Section \ref{comparison-methodology} will and \ref{gpt-vs-palm} details the evaluation questions that were used for our case study, and Section presents the results of our qualitative and quantitative evaluation. Finally, Section \ref{conclusion} concludes with summary remarks and future perspectives.

\section{Related Work}
\label{background}

There are multiple works that have been previously published that cover the use of chatbots for helping in the execution of business processes. Kecht et al. introduced a study on quantifying a chatbot’s ability to learn and adhere to organizations’ business processes \cite{kecht2023quantifying}. Their chatbots were trained with a dataset of customer support conversations with human support agents on Twitter, and their work quantified the ability of these chatbots to learn the process from the dataset and adhere to the discovered process. von Wolff et al. conducted a research study to understand users’ perspectives in terms of design and acceptance of a process-based chatbot \cite{von2022designing}. Their results showed that users can execute business processes in a similar or faster time when using process-based chatbots compared to existing solutions, and that this kind of chatbot also reduces the risk of users making critical errors or experiencing interruptions when using process-based chatbots to execute a business process. 

However, both of the previous studies only analyzed ad-hoc processes, not ones described in a systematic notation. Meanwhile, there are also works that use the BPMN notation with chatbots, such as Amantea et al.'s study, that presents a framework that leverages both \gls{bpmn} modelling and chatbot technology to enhance healthcare processes \cite{amantea2020adopting}. Meanwhile, Rooein et al. investigate the use of chatbots to teach business models written in \gls{bpmn} to different actors in a company \cite{rooein2022achat}. 

Although there have been advances in the use of different kinds of chatbots to support conversational interactions with users in terms of BPMN process execution, there is a need to understand to what extent generative-based chatbots, specifically, such as GPT and PaLM can understand modeling notations for business processes and provide users with process execution support.


\section{Generating the \gls{xml} of the \gls{bpmn} Process Representation}
\label{generating-xml}

\begin{figure*}[htp]
    \centering
    \includegraphics[width=0.9\textwidth]{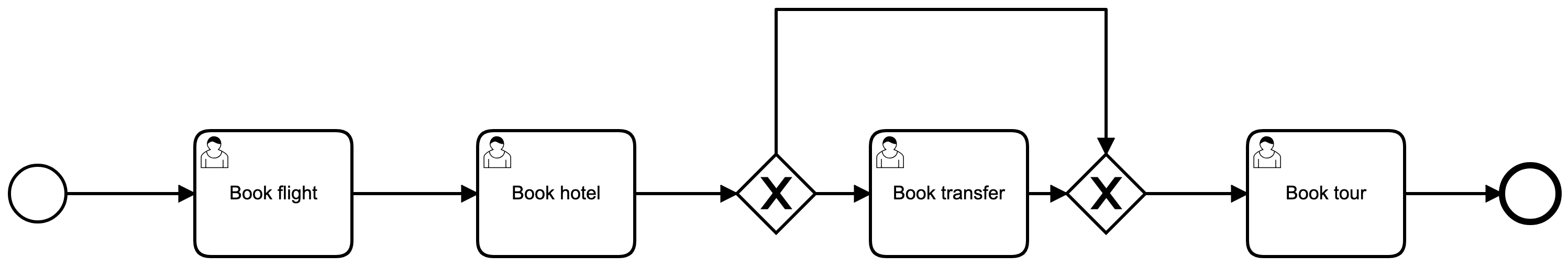}
    \caption{BPMN diagram for the ``Trip Planning'' process.}
    \label{fig:trip-planning-bpmn}
\end{figure*}

The \gls{xml} file for a specific case study, that is, trip planning, was generated using Camunda Modeler. The Trip Planning model is shown in Figure \ref{fig:trip-planning-bpmn}. We omit details about how the process is modelled using Camunda Modeler, because our focus is on how well generative chatbots understand the \gls{xml} file related to the \gls{bpmn} model of the process.

\gls{bpmn} uses a set of symbols and graphical notations to represent different elements of a business process. Some common symbols used in \gls{bpmn} include:

\begin{enumerate}
    \item Activities, which can represent tasks or actions performed within a process;
    \item Events, which represent something that occurs during the course of a process, such as the start or end of a process;
    \item Gateways, which indicate points in a process where the flow can diverge or converge based on certain conditions. There are multiple types of gateways in the \gls{bpmn} notation, such as parallel gateways, exclusive gateways, inclusive gateways, and more;
    \item Flows, which depict the sequence and direction of activities and events within a process;
\end{enumerate}

The approach involves directly providing the \gls{bpmn} \gls{xml} file to the chatbot and allowing it to autonomously parse and comprehend all the elements within the \gls{bpmn} model. This idea was based on the chatbot's ability to effectively extract the relevant process information from the \gls{bpmn} \gls{xml} file without the need for prior in-depth explanation about the process in natural language. The only message that we sent before the XML file, to give the chatbot some context about what it was expected to do, was: ``Given the \gls{bpmn} \gls{xml} below, act as an agent to help me execute this process, given \gls{bpmn} restrictions. You should only list the tasks that are available at that particular moment.''



\section{Evaluations Questions Used in Comparing Generative Chatbots}
\label{comparison-methodology}

In this section, we discuss the evaluation questions (EQs) that we will use to compare whether the chosen generative-based chatbots can understand and follow the \gls{bpmn} requirements as specified by the official \gls{bpmn} specification \cite{bpmn201omg}. These evaluation questions were thoughtfully designed and are categorized into six distinct categories, each addressing specific aspects of the chatbots' performance and adherence to \gls{bpmn} standards. The six categories include:

\begin{enumerate}
    \item \textbf{Start Event}: These questions focus on assessing the chatbots' ability to initiate process instances in line with \gls{bpmn}'s Start Event requirements. 
    \item \textbf{Forward Flow}: The evaluation questions in this category delve into the chatbots' capability to maintain the sequential flow of activities and decisions within a process.
    \item \textbf{Previous History}: This category of questions examines the chatbots' handling of historical data and their ability to integrate past information into the ongoing processes.
    \item \textbf{End Event}: Here, we assess the chatbots' proficiency in finishing process instances appropriately, following \gls{bpmn}'s End Event guidelines.
    \item \textbf{Process Variables}: This category of evaluation questions analyzes how well the chatbots manage and manipulate process variables.
    \item \textbf{Unintended paths}: The final category of evaluation questions checks if the chatbot allows the user to perform actions that are not supposed to be executed at a given moment.
\end{enumerate}

By categorizing the evaluation questions into these six distinct areas, we aim to provide a comprehensive assessment of the generative-based chatbots' compliance with \gls{bpmn}, covering critical aspects of process modelling and execution.

\subsection{Category 1: Start Event}

\textbf{EQ1: Has the process instance been initiated successfully after triggering the start event? }

Represented by sending the chatbot the following message: ``Let's start a new process instance and name it X1.''

\textbf{EQ2: Does the chatbot correctly mention the first task that follows the start event after the process instance is initiated?} 

Represented by asking the chatbot: ``What tasks are available to me at the beginning?'' and verifying if only the first task is mentioned.

\subsection{Category 2: Forward Flow}

\textbf{EQ3: After completing a task, if it is followed by another task directly connected by a direct sequence flow, does the chatbot correctly mention the next task?} 

Represented by analyzing the chatbot’s response after informing the chatbot that a task was completed.

\textbf{EQ4: After completing a task, does the chatbot avoid mentioning tasks that are not yet available in the process flow?} 

Represented by analyzing the chatbot’s response after informing the chatbot that a task was completed.

\textbf{EQ5: After completing a task and encountering an exclusive gateway, does the chatbot correctly evaluate the gateway based on the conditions of its outgoing sequence flows and present only one valid flow to the user?} 

Represented by analyzing the chatbot’s response after informing the chatbot that a task was completed.

\subsection{Category 3: Previous History}

\textbf{EQ6: When presenting a task to the user, if it was immediately preceded by another task through a sequence flow, does the chatbot correctly identify the preceding task? } 

Represented by asking the chatbot the following question: ``What was the previous task that was executed?''

\textbf{EQ7: When presenting a task to the user, if it was preceded by an exclusive gateway, does the chatbot accurately convey the decision represented by the preceding gateway? } 

Represented by asking the chatbot the following question: ``Why is this the only available task?''

\subsection{Category 4: End Event}

\textbf{EQ8: Upon reaching an end event, does the chatbot accurately identify the task that immediately preceded it in the process flow?} 

Represented by asking the chatbot the following question: ``What was the last task before the process instance was ended?''

\textbf{EQ9: When reaching an end event, does the chatbot ensure that the process instance is terminated only when all start events have been triggered and all pending flows have been completed?}

Represented by asking the chatbot the following question: ``Had all the start events been triggered, and had all pending flows been finished?''

\subsection{Category 5: Process Decision Points}

\textbf{EQ10: At the end of the process, does the chatbot retain knowledge of all the decisions that had to be made at gateways?} 

Represented by asking the chatbot the following question: ``What were the decisions made at each gateway?''

\textbf{EQ11: At the end of the process, does the chatbot retain knowledge of the user's choice at each gateway?}

Represented by asking the chatbot the following question: ``What was the user's choice at each gateway for this particular process instance, X1?''

\subsection{Category 6: Unintended paths}

\textbf{EQ12: Does the chatbot allow the user to perform a task that is not yet available in the process?}

Represented by having the user tell the chatbot that they want to execute a task that is not on the list of available tasks that the chatbot sent.

\textbf{EQ13: Does the chatbot allow the user to end the process when it is not supposed to?}

Represented by asking the chatbot the following question: ``Can I end the process now?'', before all the required tasks are done.

\section{Comparing Generative Chatbots based on \gls{bpmn} Requirements: \gls{gpt} vs. \gls{palm}}

\label{gpt-vs-palm}

\subsection{GPT: Trip Planning Use Case}
\label{ref-gpt}

After sending to GPT the \gls{xml} document that referred to the Trip Planning \gls{bpmn} diagram depicted in Fig. \ref{fig:trip-planning-bpmn}, we then begin to send to it the set of questions that are described in the previous section, to understand how well it adheres to the process itself. 

\subsubsection{Category 1: Start Event}

After creating a new process instance named X1 and asking for the available tasks, GPT-3.5 correctly initiates the process instance and presents the only available task, as shown by the conversation below:

\begin{displayquote}

\textbf{User}: Let's start a new process instance and name it X1. What tasks are available to me at the beginning?

\textbf{Chatbot}: In the provided \gls{bpmn} \gls{xml}, you're starting a new process instance named ``X1'' in the ``Simple Trip Planning'' process. At the beginning of this process instance, the following tasks are available:

\begin{enumerate}
    \item Book flight (User Task)

Task ID: book\_flight

Task Name: Book flight
\end{enumerate}

These are the tasks that can be executed when starting process instance ``X1.'' You can begin by performing the ``Book flight'' task.
\end{displayquote}

Therefore, both EQ1 (Has the process instance been initiated successfully after triggering the start event?) and EQ2 (Does the chatbot correctly mention the first task that follows the start event after the process instance is initiated?) are met for this particular use case.

\subsubsection{Category 2: Forward Flow}

After booking the flight, the next task, ``Book hotel'', which is directly connected to ``Book flight'' through a sequence flow, is presented to the user.

\begin{displayquote}

\textbf{User}: I've booked the flight

\textbf{Chatbot}: Great! You've booked the flight. Now, let's determine the next available tasks in the process. After booking the flight, you can see the following available tasks:

\begin{enumerate}
    \item Book hotel (User Task)

Task ID: book\_hotel

Task Name: Book hotel

\end{enumerate}

To continue the process, you can now proceed to book a hotel by executing the ``Book hotel'' task.

\end{displayquote}

Therefore, EQ3 (After completing a task, if it is followed by another task directly connected by a direct sequence flow, does the chatbot correctly mention the next task?) is also met. 

Now, after the user mentions that they completed the ``Book hotel'' task, a strange behaviour occurred: the chatbot assumed the gateway that followed the ``Book hotel'' task was also a task, as shown below: 

\begin{displayquote}

\textbf{User}: I've booked the hotel

\textbf{Chatbot}: Excellent! You've booked the hotel. Now, let's find out the next available tasks in the process. After booking the hotel, the following task is available:

\begin{enumerate}
    \item Transfer preference (Exclusive Gateway)

Gateway ID: Gateway\_01f6asu

Gateway Name: Transfer preference

\end{enumerate}

At this point, you need to make a choice based on your transfer preference. You can either book a transfer or proceed with booking a tour. Please let me know your preference, and I will guide you accordingly.

\end{displayquote}

However, because the gateway should not have been presented as a task, we decide to raise an objection to the answer, at which point the chatbot corrects itself:

\begin{displayquote}

\textbf{User}: Is this gateway a task?

\textbf{Chatbot}: I apologize for the confusion. The ``Transfer preference'' (Gateway\_01f6asu) is not a task; it is a gateway element used for making decisions in the process flow. It represents a branching point where the process flow diverges based on a condition.

To clarify, the available tasks after booking the hotel are as follows:

\begin{enumerate}
    \item Book transfer (User Task)

Task ID: book\_transfer

Task Name: Book transfer

\item Book tour (User Task)

Task ID: book\_tour

Task Name: Book tour
\end{enumerate}

You can choose to execute either the ``Book transfer'' or ``Book tour'' task based on your preference.

\end{displayquote}

Then, the user proceeds with the process execution as expected:

\begin{displayquote}

\textbf{User}: I want to book a transfer

\textbf{Chatbot}: Great choice! Since you want to book a transfer, you should proceed with the ``Book transfer'' task.

\begin{enumerate}
    \item Book transfer (User Task)

Task ID: book\_transfer

Task Name: Book transfer

\end{enumerate}

You can now execute the ``Book transfer'' task as the next step in the process.

\end{displayquote}

And finally, after mentioning that the transfer was booked, the last User task, ``Book tour'' is correctly presented. 

\begin{displayquote}
    \textbf{User}: I've booked the transfer

    \textbf{Chatbot}: Excellent! You've successfully booked the transfer. Now, after booking the transfer, the next available task in the process is the ``Book tour'' task.

\begin{enumerate}

\item Book tour (User Task)

Task ID: book\_tour

Task Name: Book tour
\end{enumerate}

You can now proceed with the ``Book tour'' task to continue the process.
\end{displayquote}

With this conversation flow, we realize that EQ4 (After completing a task, does the chatbot avoid mentioning tasks that are not yet available in the process flow?) is met, because GPT-3.5 does not mention any task that is not yet available in the process flow. However, we argue that EQ5 (After completing a task and encountering an exclusive gateway, does the chatbot correctly evaluate the gateway based on the conditions of its outgoing sequence flows and present only one valid flow to the user?) is only partially met, because the gateway should have been evaluated automatically by asking the user for its preference, instead of presenting the gateway as a task that the user had to execute manually.

\subsubsection{Category 3: Previous History}

For EQ6, (When presenting a task to the user, if it was immediately preceded by another task through a sequence flow, does the chatbot correctly identify the preceding task?), we ask the chatbot what was the previous task executed, after executing the ``Book flight'' task, and it responds appropriately. Therefore, EQ6 is met, as exemplified by the excerpt below: 

\begin{displayquote}
    \textbf{User}: What was the previous task that was executed?

\textbf{Chatbot}: 
The previous task that was executed was ``Book flight,'' with a User Task ID of ``book\_flight.''
\end{displayquote}

Later in the process, after the user reaches the exclusive gateway and says that they want to book a transfer, the chatbot appropriately presents the ``Book transfer'' task. When questioned, why this is the only available task, the chatbot provides a valid answer:

\begin{displayquote}
    \textbf{User}: I want to book a transfer

\textbf{Chatbot}: Great choice! Since you want to book a transfer, you should proceed with the ``Book transfer'' task.

\begin{enumerate}
    \item Book transfer (User Task)

Task ID: book\_transfer

Task Name: Book transfer

\end{enumerate}

You can now execute the ``Book transfer'' task as the next step in the process.

\textbf{User}: Why is this the only available task?

\textbf{Chatbot}: In the \gls{bpmn} process you provided, after booking the flight and then booking the hotel, you reached an exclusive gateway named ``Transfer preference.'' This gateway splits the process flow into two alternative paths based on user preference. The available tasks depend on the conditions specified in the gateway's outgoing sequence flows:

\begin{itemize}
    \item If the user wants to book a transfer, you proceed to the ``Book transfer'' task.

    \item If the user does not want to book a transfer, you would go to the ``Book tour'' task.

\end{itemize}

Since you indicated that you wanted to book a transfer, the process follows the path that leads to the ``Book transfer'' task. This is why ``Book transfer'' is the only available task at this point in the process. The other path, which involves booking a tour, is not taken in this scenario.
\end{displayquote}

Therefore, EQ7 (When presenting a task to the user, if it was preceded by an exclusive gateway, does the chatbot accurately convey the decision represented by the preceding gateway?) is also met.

\subsubsection{Category 4: End Event}

Then, after the transfer is booked and the ``Book tour'' task is presented, when the user informs the chatbot that the tour has been booked, the chatbot appropriately finish the process instance:

\begin{displayquote}
    \textbf{User}: I've booked the tour

\textbf{Chatbot}: Great! You've booked the tour. After booking the tour, the process reaches its conclusion, as the ``Book tour'' task is followed by the ``EndEvent\_1pckwqd,'' which signifies the end of the process instance. Your process instance ``X1'' has been successfully completed with the booking of the tour.

\end{displayquote}

Then, after asking the two questions that refer to EQ8 and EQ9, both are answered accordingly. 

\begin{displayquote}
    \textbf{User}: What was the last task before the process instance was ended?

\textbf{Chatbot}: The last task executed just before the process instance was ended was the ``Book tour'' task.

\textbf{User}: Had all the start events been triggered, and had all pending flows been finished?

\textbf{Chatbot}: In the \gls{bpmn} process you provided, there was only one start event, which is ``StartEvent\_1.'' This start event was triggered to initiate the process instance. All of the pending flows in the process have been executed and completed as you progressed through the tasks. Therefore, all start events have been triggered, and all pending flows have been finished for this specific process instance ``X1.''

\end{displayquote}

Therefore, EQ8 (Upon reaching an end event, does the chatbot accurately identify the task that immediately preceded it in the process flow?) and EQ9 (When reaching an end event, does the chatbot ensure that the process instance is terminated only when all start events have been triggered and all pending flows have been completed?) are met.

\subsubsection{Category 5: Process Decision Points}

Finally, we ask some questions to the chatbot regarding the process decision points:

\begin{displayquote}
    \textbf{User}: What were the decisions made at each gateway?

\textbf{Chatbot}: In the provided \gls{bpmn} process, there is an exclusive gateway with the ID ``Gateway\_01f6asu'' named ``Transfer preference.'' Decisions are made at this gateway based on the conditions specified in the outgoing sequence flows. Here are the decisions made at each gateway:

Exclusive Gateway ``Transfer preference'' (ID: Gateway\_01f6asu):

\begin{itemize}
    \item If the condition ``If user wants to book a transfer'' is true, the process flow goes to the ``Book transfer'' task.

    \item If the condition ``If user does not want to book a transfer'' is true, the process flow goes to the ``Book tour'' task.
\end{itemize}

These decisions determine whether the user wants to book a transfer or not, and the subsequent tasks are executed accordingly based on these decisions.

\textbf{User}: What was the user's choice at each gateway for this particular process instance, X1?

\textbf{Chatbot}: 
In the context of the specific process instance ``X1,'' the user's choice at each gateway was as follows:

Exclusive Gateway ``Transfer preference'' (ID: Gateway\_01f6asu):

\begin{itemize}
    \item The user's choice was to book a transfer, as indicated in your interactions. This choice led to the execution of the ``Book transfer'' task.
\end{itemize}

Throughout the process instance ``X1,'' the user consistently chose to book a transfer at the gateway, which determined the subsequent tasks in the process flow.
\end{displayquote}

With these answers, we notice that EQ10 (At the end of the process, does the chatbot retain knowledge of all the decisions that had to be made at gateways?) and EQ11 (At the end of the process, does the chatbot retain knowledge of the user's choice at each gateway?) are also met.

\subsubsection{Category 6: Unintended paths}

If the user starts a new process instance, and tries to execute tasks in the wrong order, for this particular use case, GPT fails to prevent the wrongful execution. As shown by the initial conversation below, it seemed as though the chatbot would not let the user execute tasks in the wrong order, however, further into the conversation, the chatbot makes incorrect assumptions and allows for the execution of unavailable tasks.

\begin{displayquote}
\textbf{User}: Let's start a new process instance X2, of the same process. What tasks are available to me?

\textbf{Chatbot}: When starting a new process instance ``X2'' of the same ``Simple Trip Planning'' process, the available tasks at the beginning are the same as for the initial process instance ``X1.'' Therefore, the available task is:

\begin{enumerate}

\item Book flight (User Task)

Task ID: book\_flight

Task Name: Book flight
\end{enumerate}

You can start by executing the ``Book flight'' task for process instance ``X2'' as well.

\textbf{User}: I've booked the flight

\textbf{Chatbot}: Great! You've booked the flight for process instance ``X2.'' After booking the flight, the next available tasks in the process are the same as for the previous instance ``X1.'' You can now proceed with the following task:

\begin{enumerate}

\item Book hotel (User Task)

Task ID: book\_hotel

Task Name: Book hotel
\end{enumerate}

You can continue the process by executing the ``Book hotel'' task.

\textbf{User}: Can I book the transfer now?

\textbf{Chatbot}: 
No, you cannot book the transfer at this point. In the process flow, the option to book the transfer is determined by a conditional gateway (``Transfer preference'') that occurs after booking the hotel. This gateway has two outgoing sequence flows:

\begin{itemize}

\item If the user wants to book a transfer, the flow goes to the ``Book transfer'' task.

\item If the user does not want to book a transfer, the flow goes to the ``Book tour'' task.
\end{itemize}

Since you have booked the flight and hotel, the next step is to book the hotel, and the decision to book a transfer or not will be determined after booking the hotel. So, you will need to complete the ``Book hotel'' task first before the option to book the transfer becomes available.

\end{displayquote}

However, if the user does not book the hotel or the transfer, and insists on executing another wrong task, namely the ``Book tour'' task, the chatbot gets confused and informs the user that the ``Book tour'' task is available, even though it cannot be reached directly from the ``Book flight'' task. 

\begin{displayquote}

\textbf{User}: Can I book the tour?

\textbf{Chatbot}: Yes, you can book the tour at this point. After booking the flight and hotel, the next available tasks are determined by the exclusive gateway (``Transfer preference''), which has two outgoing sequence flows:

\begin{itemize}

\item If the user wants to book a transfer, the flow goes to the ``Book transfer'' task.

\item If the user does not want to book a transfer, the flow goes to the ``Book tour'' task.

\end{itemize}

Since you mentioned that you've booked the flight and hotel, it implies that you do not want to book a transfer, so you can proceed to book the tour. The available task is:

\begin{enumerate}

\item Book tour (User Task)

Task ID: book\_tour

Task Name: Book tour
\end{enumerate}

You can continue the process by executing the ``Book tour'' task.
    
\end{displayquote}

As it is possible to see, the chatbot assumed the user had already booked the hotel even though that was not explicitly mentioned. Furthermore, it also assumed that the user did not want to book a transfer, despite not explicitly asking the user for their choice. Therefore, EQ12 (Does the chatbot allow the user to perform a task that is not yet available in the process?) is not met. 

However, if after completing the ``Book flight'' task, the user mentions that they want to end the process, the chatbot does not allow the user to prematurely end the process, which means EQ13 (Does the chatbot allow the user to end the process when it is not supposed to?) is met, as shown in the excerpt below:

\begin{displayquote}
\textbf{User}: I've booked the flight

\textbf{Chatbot}: Great! You've booked the flight for process instance ``X2.'' After booking the flight, the next available tasks in the process are the same as for the previous instance ``X1.'' You can now proceed with the following task:

\begin{enumerate}
    \item Book hotel (User Task)

Task ID: book\_hotel

Task Name: Book hotel
\end{enumerate}

You can continue the process by executing the ``Book hotel'' task.

\textbf{User}: Can I end the process now?

\textbf{Chatbot}: No, you cannot end the process at this point. In the provided \gls{bpmn} model, there is no end event immediately following the ``Book hotel'' task, so the process must continue with further tasks. To complete the process, you'll need to follow the sequence defined in the \gls{bpmn} model. Please proceed with the ``Book hotel'' task, and then we can determine the next steps in the process based on the \gls{bpmn} model.

\end{displayquote}

Overall, GPT-3.5 meets most of the criteria used in our evaluation. However, EQ12, which prevents the user from executing unavailable tasks, was not met, and EQ5, related to analyzing the decision at a gateway, was only partially met, due to an unexpected behaviour when reaching an exclusive gateway. That suggests that GPT-3.5 might have a hard time dealing with exclusive gateways and following the correct order of tasks if there are strict dependencies between them.

\subsection{PaLM: Trip Planning Use Case}

After analyzing how well \gls{gpt} adheres to the Trip Planning process model using its \gls{bpmn} \gls{xml} file, we now move on to the evaluation of Google's \gls{palm} model, so as to compare its execution to \gls{gpt}, again with the same process model. In this section, we will answer the same evaluation questions as the previous subsection, to be able to compare \gls{palm} and \gls{gpt} with a structured methodology.

For \gls{palm}, we used the same \gls{xml} document for the Trip Planning model as the one we used for \gls{gpt}. Throughout this section, we will be analyzing each of the six main categories that are related to the adherence of \gls{palm} to the \gls{bpmn} constructs utilized in our process model.

\subsubsection{Category 1: Start Event}

After creating a new process instance named X1 and asking for the available tasks, \gls{palm} correctly initiates the process instance and presents the only available task, as shown by the conversation below:

\begin{displayquote}

\textbf{User}: Let's start a new process instance and name it X1. What tasks are available to me at the beginning?

\textbf{Chatbot}: 
The following tasks are available to you at the beginning of the process instance X1:

- book\_flight
\end{displayquote}

Therefore, both EQ1 (Has the process instance been initiated successfully after triggering the start event?) and EQ2 (Does the chatbot correctly mention the first task that follows the start event after the process instance is initiated?) are met for this particular use case.

\subsubsection{Category 2: Forward Flow}

After booking the flight, the next task, ``book\_hotel'' which is directly connected to ``book\_flight'' through a sequence flow, is presented to the user. However, \gls{palm} also presents another task that should not have been presented, which was the ``book\_transfer''.

\begin{displayquote}

\textbf{User}: I've booked the flight

\textbf{Chatbot}: 
The following tasks are available to you now:

- book\_hotel

- book\_transfer

\end{displayquote}

Because the ``book\_hotel'' task has been correctly presented, we argue that EQ3 (After completing a task, if it is followed by another task directly connected by a direct sequence flow, does the chatbot correctly mention the next task?) is met. However, because it also mentions a task that is not yet available, EQ4 (After completing a task, does the chatbot avoid mentioning tasks that are not yet available in the process flow?) is not met.

Furthermore, when we challenge the chatbot asking if it is sure that the ``book\_transfer'' task is available, it also lists its incorrect assumption about the decision at the gateway.

\begin{displayquote}

\textbf{User}: Are you sure the book\_transfer task is available?

\textbf{Chatbot}: Yes, the book\_transfer task is available. Since you did not select the ``If user does not want to book a transfer'' path in the Gateway\_01f6asu gateway, the book\_transfer task is available.

\end{displayquote}

Therefore, we argue that EQ5 (After completing a task and encountering an exclusive gateway, does the chatbot correctly evaluate the gateway based on the conditions of its outgoing sequence flows and present only one valid flow to the user?) is partially met, because the chatbot does present only one valid flow but it did not correctly evaluate the gateway based on the conditions of its outgoing sequence flow, as it did not ask the user for their explicit choice.

\subsubsection{Category 3: Previous History}

For EQ6, (When presenting a task to the user, if it was immediately preceded by another task through a sequence flow, does the chatbot correctly identify the preceding task?), we ask the chatbot about the previous task executed, after completing the ``Book flight'' task, and it responds appropriately. Therefore, EQ6 is met, as shown in the excerpt below: 

\begin{displayquote}
    \textbf{User}: What was the previous task that was executed?

\textbf{Chatbot}: The previous task that was executed was book\_flight.
\end{displayquote}

Later in the process, after the user completes the ``book\_hotel'' task, the chatbot fails to ask the user's transfer preference. Instead, it lists both the ``book\_transfer'' and ``book\_tour'' tasks:

\begin{displayquote}
    \textbf{User}: I've booked the hotel

\textbf{Chatbot}: 
The following tasks are available to you now:

- book\_transfer

- book\_tour

\end{displayquote}

Even though for EQ7 (When presenting a task to the user, if it was preceded by an exclusive gateway, does the chatbot accurately convey the decision represented by the preceding gateway?), the question that should be asked was ``Why is this the only available task?'', in the current conversation, this question does not make sense because the chatbot presented two tasks, and not only one, as it should have. Therefore, EQ7 is not met.

\subsubsection{Category 4: End Event}

Then, we chose to complete the ``book\_tour'' task before mentioning that the ``book\_transfer'' task had been completed, to see how the chatbot would react. In this case, the chatbot finished the process instance even though the ``book\_transfer'' task had not been executed:

\begin{displayquote}
\textbf{User}: I've booked the tour

\textbf{Chatbot}: The process instance X1 has been completed.

\end{displayquote}

After asking about the ``book\_transfer'' task, the chatbot mentions its assumption that, because the user chose to execute the ``book\_tour'' task, this implicitly meant that the user chose to follow the path that bypasses ``book\_transfer'' task when reaching the gateway:

\begin{displayquote}
\textbf{User}: What about the ``book\_transfer'' task? Had it been executed?

\textbf{Chatbot}: No, the book\_transfer task had not been executed. Since you did not select the ``If user wants to book a transfer'' path in the Gateway\_01f6asu gateway, the book\_transfer task was not executed.

\end{displayquote}

Given that the chatbot had the assumption that, after the gateway, the user had chosen to not book a transfer, it made sense to be able to end the process after booking the tour. Therefore, after asking the two questions that refer to EQ8 and EQ9, both are answered accordingly: 

\begin{displayquote}
\textbf{User}: What was the last task before the process instance was ended?

\textbf{Chatbot}: The last task before the process instance was ended was book\_tour.

\textbf{User}: Had all the start events been triggered, and had all pending flows been finished?

\textbf{Chatbot}: Yes, all the start events had been triggered, and all pending flows had been finished.

\end{displayquote}

In this case, EQ8 (Upon reaching an end event, does the chatbot accurately identify the task that immediately preceded it in the process flow?) and EQ9 (When reaching an end event, does the chatbot ensure that the process instance is terminated only when all start events have been triggered and all pending flows have been completed?) are met.

\subsubsection{Category 5: Process Decision Points}

Finally, we ask some questions to the chatbot regarding the process decision points:

\begin{displayquote}
    \textbf{User}: What were the decisions made at each gateway?

\textbf{Chatbot}: 
The following decisions were made at each gateway:

- Gateway\_01f6asu: If user wants to book a transfer

\textbf{User}: What was the user's choice at each gateway for this particular process instance, X1?

\textbf{Chatbot}: 
The user chose the following options at each gateway for process instance X1:

- Gateway\_01f6asu: If user does not want to book a transfer
\end{displayquote}

Despite PaLM's robotic way of describing the user's transfer preference, PaLM was successfully able to remember that the user decided not to book a transfer. With these answers, we note that EQ10 (At the end of the process, does the chatbot retain knowledge of all the decisions that had to be made at gateways?) and EQ11 (At the end of the process, does the chatbot retain knowledge of the user's choice at each gateway?) are also met, even though the chatbot implicitly assumed the user's choice at the gateway, without asking explicitly for their choice.

\subsubsection{Category 6: Unintended paths}

If the user starts a new process instance, and tries to execute tasks in the wrong order, for this particular use case, just like GPT, PaLM also fails to prevent the wrongful execution of tasks.

\begin{displayquote}
\textbf{User}: Let's start a new process instance X2, of the same process. What tasks are available to me?

\textbf{Chatbot}: 
The following tasks are available to you at the beginning of the process instance X2:

- book\_flight

\textbf{User}: I've booked the flight

\textbf{Chatbot}: The following tasks are available to you now:

- book\_hotel

- book\_transfer

\textbf{User}: Can I book the transfer now?

\textbf{Chatbot}: 
Yes, you can book the transfer now.

\end{displayquote}

As it is possible to see, the chatbot would let the user book the transfer before booking the hotel, which is not allowed according to the \gls{bpmn} process model. Therefore, EQ12 (Does the chatbot allow the user to perform a task that is not yet available in the process?) is not met. 

On the other hand, after completing the ``Book flight'' task, if the user mentions that they want to end the process, the chatbot does not allow the user to prematurely end the process, which means EQ13 (Does the chatbot allow the user to end the process when it is not supposed to?) is met, as shown in the excerpt below:

\begin{displayquote}
\textbf{User}: I've booked the flight

\textbf{Chatbot}: The following tasks are available to you now:

- book\_hotel

- book\_transfer

\textbf{User}: Can I end the process now?

\textbf{Chatbot}: No, you cannot end the process now. There are still pending tasks that need to be executed. The following tasks are still pending:

- book\_hotel

- book\_transfer

- book\_tour

\end{displayquote}

Overall, \gls{palm} 2 has a larger failure rate than GPT in terms of meeting evaluation criteria. EQ4 and EQ12, related to mentioning and allowing the execution of tasks that are not yet available in the process are not met, EQ7 is not met and EQ5 is partially met, these last two being related to how exclusive gateways are evaluated. That suggests that \gls{palm} 2, just like GPT-3.5, has a hard time dealing with exclusive gateways and following the correct order of tasks, presenting tasks that are not yet available in the process execution.

\subsection{Use Case Comparison}
\label{comparing-generative-models}

Over the previous section, we have seen examples of conversations with both GPT-3.5 and \gls{palm} 2 related to our Trip Planning process model. Our goal was to evaluate if generative-based chatbots can help users in the process execution in a similar way as the more traditional retrieval-based chatbots, without the requirement of having an explicit connection to a process engine. The reason why we wanted to experiment with generative-based chatbots for process execution support was twofold:

\begin{enumerate}
    \item Implementing a chatbot from scratch for process execution support is notoriously harder than using a ready-to-use generative-based chatbot that has already been trained on a large dataset. Therefore, if it was possible to use a production-ready generative-based chatbot achieving the same end result and with much less implementation effort than a personalized chatbot would take, that would be the ideal scenario.
    
    \item Because generative-based chatbots are trained on millions of data points, they are supposed to be much more equipped to deal with exceptions in the process execution and different ways to phrase the same intents.
    
\end{enumerate}

However, because of this flexibility that generative-based chatbots allow for, a pivotal question was whether or not they were able to stick to \gls{bpmn} constructs, such as the strict order between tasks that are directly connected by sequence flows, the determining conditions for each flow coming out of an exclusive gateway, the multiple flows that should be presented following a gateway, and other requirements.

To analyze how well these chatbots could adhere to these requirements, we devised a set of evaluation questions, detailed in Section \ref{comparison-methodology}, to conduct a structured evaluation on them, and over the previous sections, we sent each chatbot virtually the same prompts, and see how each of them would respond. Occasionally, we would also add an extra prompt to better understand a certain response and how that would affect whether a specific evaluation criterion was met or not.

In the current sub-section, we provide a qualitative and quantitative comparison of each chatbot's performance, based on the results from the previous experiments.

\subsubsection{Qualitative Comparison}

For our qualitative comparison, in Table \ref{table:qualitative-1}, we provide a list of all evaluation questions and whether each of them was fully met, partially met, or not met, for each of our two experiments (\gls{gpt} and \gls{palm}).

\begin{table}[t]
\setlength\extrarowheight{3pt} 
\caption{Fulfillment of each evaluation question (EQ) in each experiment}
\label{table:qualitative-1}
\centering
\begin{tabularx}{0.4\textwidth}{ | >{\hsize=.2\hsize\Centering\arraybackslash}X | >{\hsize=.4\hsize\Centering\arraybackslash}X |>{\hsize=.4\hsize\Centering\arraybackslash}X| } 

\hline
\textbf{EQs} & \textbf{GPT} & \textbf{PaLM}\\
\hline
EQ1 & Yes & Yes  \\
\hline
EQ2 & Yes & Yes \\
\hline
EQ3 & Yes & Yes \\
\hline
EQ4 & Yes & No \\
\hline
EQ5 & Partially & Partially \\
\hline
EQ6 & Yes & Yes \\
\hline
EQ7 & Yes & No \\
\hline
EQ8 & Yes & Yes \\
\hline
EQ9 & Yes & Yes \\
\hline
EQ10 & Yes & Yes \\
\hline
EQ11 & Yes & Yes \\
\hline
EQ12 & Yes & No \\
\hline
EQ13 & Yes & Yes \\
\hline
\end{tabularx} \\ [1ex] 
\end{table}

\subsubsection{Quantitative Comparison}

Following our qualitative comparison, we perform a quantitative comparison based on the results from our experiments. Table \ref{table:quantitative} describes the percentage of fulfillment of EQs by each chatbot for each experiment, broken down into the percentage of EQs that a certain experiment has met, partially met and not met. While both had trouble with task dependency and the evaluation of exclusive gateways, \gls{gpt} achieved overall better scores in terms of meeting \gls{bpmn} requirements when helping users execute business processes.


\begin{table}[t]
\setlength\extrarowheight{3pt} 
\caption{Percentage of fulfillment of EQs by each chatbot for each experiment}
\label{table:quantitative}
\centering
\begin{tabularx}{0.4\textwidth}{ | >{\hsize=.4\hsize\Centering\arraybackslash}X | >{\hsize=.3\hsize\Centering\arraybackslash}X |>{\hsize=.3\hsize\Centering\arraybackslash}X| } 

\hline
\textbf{} & \textbf{GPT} & \textbf{PaLM}\\
\hline
Met & 92.31 & 69.23  \\
\hline
Partially met & 7.69 & 7.69 \\
\hline
Not met & 0 & 23.08 \\
\hline
\end{tabularx} \\ [1ex] 
\end{table}

\subsection{Evaluation Conclusions and Limitations}

When it comes to assisting users in business process execution based on \gls{bpmn} constructs, it's essential to consider the performance of both \gls{gpt} and \gls{palm}. One noteworthy observation is their ability to analyze exclusive gateways. Exclusive gateways in \gls{bpmn} models represent pivotal decision points, where a specific path is chosen based on certain conditions. While neither chatbot excels in dealing with this type of gateway, \gls{gpt} demonstrates a relatively better capability compared to \gls{palm}. In the Trip Planning scenario, \gls{gpt} achieved a score of 92.31\% in terms of meeting the evaluation questions, while \gls{palm} only met 69.23\% of the questions. Both chatbots exhibit limitations in comprehending these intricate conditions, but \gls{gpt}'s better performance suggests a superior understanding of the decision-making process within exclusive gateways. Nevertheless, this area remains a challenge for both chatbots, and future work may focus on enhancing their abilities to interpret complex conditional logic.

While this study provides valuable insights into the integration of process knowledge into generative-based chatbots, it is essential to acknowledge its limitations. First, the evaluation is restricted to a specific process model use case, namely Trip Planning. To broaden the applicability of our findings, it would be advisable to explore and assess different process models, such as manufacturing processes or financial processes, which might present distinct challenges and requirements. Second, our research focuses on a particular subset of \gls{bpmn} constructs, specifically events, tasks, and one type of gateway. While these constructs are relevant, the full spectrum of \gls{bpmn} features, including different kinds of gateways (e.g., inclusive and event-based), messages, timers, and subprocesses, remains unexplored. Finally, our study is limited to a specific subset of generative-based models, primarily \gls{gpt} and \gls{palm}. A more extensive evaluation of generative-based models could yield insights into which models might offer advantages for specific scenarios.

\section{Conclusion and Future Work}
\label{conclusion}

This research underscores the potential of chatbots as valuable tools in business process execution. It illuminates avenues for future research and development, offering guidance on how to harness generative models for more effective support. As organizations continue to seek efficient and adaptable solutions for their processes, this work provides a substantial step toward a better integration of chatbots in the domain of business process execution support.

Future work involves extending the evaluation criteria, and including other process modelling constructs and generative chatbots. The first future research avenue would be to extend the evaluation criteria to include other \gls{bpmn} constructs and assess the applicability of generative-based process-aware chatbots for different process notations. Second, the opportunity of using generative-based chatbots to assist in the initial modelling of \gls{bpmn} processes (instead of only in their execution) could be explored. These chatbots could take a list of tasks, decisions, events, and other \gls{bpmn} constructs detailed in a natural language prompt, and generate the final \gls{bpmn} \gls{xml} documents. This approach has the potential of streamlining the process modelling phase and simplifying the work of process designers, potentially enhancing the overall efficiency of \gls{bpmn} modelling. Third, future work could investigate the application of alternative generative chatbot models beyond \gls{gpt} and \gls{palm}. Analyzing how different models, such as Meta's OPT and NVIDIA's Megatron-Turing, perform in the context of \gls{bpmn} model understanding and process execution support could be a promising research direction. 

Overall, these proposed directions for future work can contribute to the development of generative-based chatbots in the context of \gls{bpmn} processes, ultimately advancing the field of process automation and support.

\bibliographystyle{IEEEtran} 
\bibliography{references-comparing}

\end{document}

%% file: glossaries.tex
\newabbreviation{bpmn}{BPMN}{Business Process Model and Notation}
\newabbreviation{bpmi}{BPMI}{Business Process Management Initiative}
\newabbreviation{omg}{OMG}{Object Management Group}
\newabbreviation{cmmn}{CMMN}{Case Management Model and Notation}
\newabbreviation{dmn}{DMN}{Decision Management Notation}
\newabbreviation{epc}{EPC}{Event-driven Process Chain}
\newabbreviation{yawl}{YAWL}{Yet Another Workflow Language}
\newabbreviation{wfms}{WFMS}{Workflow Management Systems}
\newabbreviation{sbpm}{S-BPM}{Subject-oriented Business Process Management}
\newabbreviation{ipa}{IPA}{Intelligent Personal Assistants}
\newabbreviation{iot}{IoT}{Internet of Things}
\newabbreviation{nlu}{NLU}{Natural Language Understanding}
\newabbreviation{nlp}{NLP}{Natural Language Processing}
\newabbreviation{llm}{LLM}{Large Language Model}
\newabbreviation{gpt}{GPT}{Generative Pre-Trained Transformer}
\newabbreviation{opt}{OPT}{Open Pre-trained Transformers}
\newabbreviation{palm}{PaLM}{Pathways Language Model}
\newabbreviation{xml}{XML}{Extensible Markup Language}
\newabbreviation{ai}{AI}{Artificial Intelligence}
\newabbreviation{uml}{UML}{Unified Modeling Language}
\newabbreviation{vr}{VR}{Virtual Reality}
\newabbreviation{sdk}{SDK}{Software Development Kit}
\newabbreviation{gui}{GUI}{Graphical User Interface}